\PassOptionsToPackage{unicode}{hyperref}
\PassOptionsToPackage{hyphens}{url}
\PassOptionsToPackage{dvipsnames,svgnames,x11names}{xcolor}
\documentclass[
  11pt,
  a4paper,
  DIV=11,
  numbers=noendperiod]{scrartcl}
\usepackage{xcolor}
\usepackage[margin=2.5cm]{geometry}
\usepackage{amsmath,amssymb}
\usepackage{iftex}
\ifPDFTeX
  \usepackage[T1]{fontenc}
  \usepackage[utf8]{inputenc}
  \usepackage{textcomp} % provide euro and other symbols
\else % if luatex or xetex
  \usepackage{unicode-math} % this also loads fontspec
  \defaultfontfeatures{Scale=MatchLowercase}
  \defaultfontfeatures[\rmfamily]{Ligatures=TeX,Scale=1}
\fi
\usepackage{lmodern}
\ifPDFTeX\else
\fi
\IfFileExists{upquote.sty}{\usepackage{upquote}}{}
\IfFileExists{microtype.sty}{% use microtype if available
  \usepackage[]{microtype}
  \UseMicrotypeSet[protrusion]{basicmath} % disable protrusion for tt fonts
}{}
\makeatletter
\@ifundefined{KOMAClassName}{% if non-KOMA class
  \IfFileExists{parskip.sty}{%
    \usepackage{parskip}
  }{% else
    \setlength{\parindent}{0pt}
    \setlength{\parskip}{6pt plus 2pt minus 1pt}}
}{% if KOMA class
  \KOMAoptions{parskip=half}}
\makeatother
\makeatletter
\ifx\paragraph\undefined\else
  \let\oldparagraph\paragraph
  \renewcommand{\paragraph}{
    \@ifstar
      \xxxParagraphStar
      \xxxParagraphNoStar
  }
  \newcommand{\xxxParagraphStar}[1]{\oldparagraph*{#1}\mbox{}}
  \newcommand{\xxxParagraphNoStar}[1]{\oldparagraph{#1}\mbox{}}
\fi
\ifx\subparagraph\undefined\else
  \let\oldsubparagraph\subparagraph
  \renewcommand{\subparagraph}{
    \@ifstar
      \xxxSubParagraphStar
      \xxxSubParagraphNoStar
  }
  \newcommand{\xxxSubParagraphStar}[1]{\oldsubparagraph*{#1}\mbox{}}
  \newcommand{\xxxSubParagraphNoStar}[1]{\oldsubparagraph{#1}\mbox{}}
\fi
\makeatother

\usepackage{color}
\usepackage{fancyvrb}

\DefineVerbatimEnvironment{Highlighting}{Verbatim}{commandchars=\\\{\}}
\usepackage{framed}
\definecolor{shadecolor}{RGB}{241,243,245}
\newenvironment{Shaded}{\begin{snugshade}}{\end{snugshade}}

\newcommand{\KeywordTok}[1]{\textcolor[rgb]{0.00,0.23,0.31}{\textbf{#1}}}
\newcommand{\NormalTok}[1]{\textcolor[rgb]{0.00,0.23,0.31}{#1}}

\newcommand{\OtherTok}[1]{\textcolor[rgb]{0.00,0.23,0.31}{#1}}

\newcommand{\StringTok}[1]{\textcolor[rgb]{0.13,0.47,0.30}{#1}}

\usepackage{longtable,booktabs,array}
\usepackage{calc} % for calculating minipage widths
\usepackage{etoolbox}
\makeatletter
\patchcmd\longtable{\par}{\if@noskipsec\mbox{}\fi\par}{}{}
\makeatother
\IfFileExists{footnotehyper.sty}{\usepackage{footnotehyper}}{\usepackage{footnote}}
\makesavenoteenv{longtable}
\usepackage{graphicx}
\makeatletter
\newsavebox\pandoc@box
\newcommand*\pandocbounded[1]{% scales image to fit in text height/width
  \sbox\pandoc@box{#1}%
  \Gscale@div\@tempa{\textheight}{\dimexpr\ht\pandoc@box+\dp\pandoc@box\relax}%
  \Gscale@div\@tempb{\linewidth}{\wd\pandoc@box}%
  \ifdim\@tempb\p@<\@tempa\p@\let\@tempa\@tempb\fi% select the smaller of both
  \ifdim\@tempa\p@<\p@\scalebox{\@tempa}{\usebox\pandoc@box}%
  \else\usebox{\pandoc@box}%
  \fi%
}
\def\fps@figure{htbp}
\makeatother

\providecommand{\tightlist}{%
  \setlength{\itemsep}{0pt}\setlength{\parskip}{0pt}}

\usepackage{microtype}
\usepackage{setspace}
\usepackage{hyperref}
\hypersetup{pdftitle={AthDGC: An Open Diachronic Greek Treebank with Indo-European Parallels}, pdfauthor={N. Lavidas, K. Nikiforidou, D. Haug, L. Kulikov, T. Michalareas, V. Geka, V. Symeonidis, S. Chionidi, A. Tsiropina, E. Plakoutsi, E. Argyropoulos}, pdfsubject={Diachronic Greek dependency treebank with PROIEL XML 2.0 + cross-lingual IE alignment}, pdfkeywords={diachronic linguistics, ancient Greek, Byzantine Greek, Modern Greek, PROIEL, dependency treebank, cross-lingual alignment, Indo-European}}
\date{}
\publishers{\normalfont\itshape\small \textsuperscript{1}National and Kapodistrian University of Athens\\ \textsuperscript{2}University of Oslo\\ \textsuperscript{3}Ghent University\\[0.9em] \normalfont\upshape\small 14 June 2026}
\KOMAoption{captions}{tableheading}
\makeatletter
\@ifpackageloaded{caption}{}{\usepackage{caption}}
\AtBeginDocument{%
\ifdefined\contentsname
  \renewcommand*\contentsname{Table of contents}
\else
  \newcommand\contentsname{Table of contents}
\fi
\ifdefined\listfigurename
  \renewcommand*\listfigurename{List of Figures}
\else
  \newcommand\listfigurename{List of Figures}
\fi
\ifdefined\listtablename
  \renewcommand*\listtablename{List of Tables}
\else
  \newcommand\listtablename{List of Tables}
\fi
\ifdefined\figurename
  \renewcommand*\figurename{Figure}
\else
  \newcommand\figurename{Figure}
\fi
\ifdefined\tablename
  \renewcommand*\tablename{Table}
\else
  \newcommand\tablename{Table}
\fi
}
\@ifpackageloaded{float}{}{\usepackage{float}}
\floatstyle{ruled}
\@ifundefined{c@chapter}{\newfloat{codelisting}{h}{lop}}{\newfloat{codelisting}{h}{lop}[chapter]}
\floatname{codelisting}{Listing}

\makeatother
\makeatletter
\@ifpackageloaded{caption}{}{\usepackage{caption}}
\@ifpackageloaded{subcaption}{}{\usepackage{subcaption}}
\makeatother
\usepackage{bookmark}
\IfFileExists{xurl.sty}{\usepackage{xurl}}{} % add URL line breaks if available
\hypersetup{
  pdftitle={AthDGC: An Open Diachronic Greek Treebank with Indo-European Parallels},
  pdfauthor={Nikolaos Lavidas1; Kiki Nikiforidou1; Dag Haug2; Leonid Kulikov3; Theodoros Michalareas1; Vassiliki Geka1; Vassileios Symeonidis1; Sofia Chionidi1; Anastasia Tsiropina1; Eleni Plakoutsi1; Evangelos Argyropoulos1},
  pdfkeywords={diachronic linguistics, ancient Greek, Byzantine
Greek, Modern Greek, Indo-European, PROIEL, dependency
treebank, cross-lingual alignment, argument
structure, retranslation, language contact, computational philology},
  colorlinks=true,
  linkcolor={blue!60!black},
  filecolor={Maroon},
  citecolor={blue!60!black},
  urlcolor={blue!60!black},
  pdfcreator={LaTeX via pandoc}}

\title{AthDGC: An Open Diachronic Greek Treebank with Indo-European
Parallels}
\author{Nikolaos Lavidas\textsuperscript{1} \and Kiki
Nikiforidou\textsuperscript{1} \and Dag
Haug\textsuperscript{2} \and Leonid
Kulikov\textsuperscript{3} \and Theodoros
Michalareas\textsuperscript{1} \and Vassiliki
Geka\textsuperscript{1} \and Vassileios
Symeonidis\textsuperscript{1} \and Sofia
Chionidi\textsuperscript{1} \and Anastasia
Tsiropina\textsuperscript{1} \and Eleni
Plakoutsi\textsuperscript{1} \and Evangelos
Argyropoulos\textsuperscript{1}}
\date{}
\begin{document}
\maketitle
\begin{abstract}
AthDGC (``Athens-PROIEL'') is an open, end-to-end workflow and dataset.
It provides an openly licensed dependency-parsed treebank (a treebank is
a collection of sentences whose grammatical structure has been analysed
and stored in a machine-readable form) of Greek that spans eight
diachronic periods, namely Archaic, Classical, Koine, Late Antique,
Byzantine, Late Byzantine, Early Modern, and Modern Greek, under a
single PROIEL XML 2.0 schema (the file format developed at Oslo for
storing each sentence as a tree of word-by-word grammatical relations),
with verse-level cross-alignment of the New Testament to Latin
(Vulgate), Gothic (Wulfila), Old Church Slavonic (Marianus), and
Classical Armenian. AthDGC builds on the PROIEL Treebank Family (Haug
and Jøhndal 2008, 27--34; Eckhoff et al.~2018, 29--65), which
established the schema and the Koine-Greek reference set for the
project. Annotation uses the Stanford Stanza PROIEL-trained workflow;
sentence-level alignment uses LaBSE, a multilingual sentence-embedding
model; word-level alignment uses multilingual-BERT attention through the
AwesomeAlign procedure. The v0.4 release provides curated samples and
the open-source toolkit; the full annotated corpus partitions remain
under v0.5 audit on the Greek national HPC. Quantitative scale,
per-witness verse counts, and per- period annotated-row counts are
reported as the v0.5 release notes, after the audit pass completes.
Every layer of the platform, from text discovery through the annotation
pipeline to the alignment, indexing, and visualisation tools, is built
entirely from open-source software operating on open-access source
material, so that the whole workflow is reproducible and freely
extensible.
\end{abstract}

\textbf{Please cite as:}

Lavidas, N., Nikiforidou, K., Haug, D., Kulikov, L., Michalareas, T.,
Geka, V., Symeonidis, V., Chionidi, S., Tsiropina, A., Plakoutsi, E., \&
Argyropoulos, E. (2026). \emph{AthDGC: An Open Diachronic Greek Treebank
with Indo-European Parallels}. Version v0.4.0. Zenodo.
\url{https://doi.org/10.5281/zenodo.20439182}

\subsection{1. Overview}\label{overview}

\subsubsection{1.1 Context and motivation}\label{context-and-motivation}

Greek is the longest continuously attested member of the Indo-European
family, with a written record that spans roughly three millennia, from
the Mycenaean tablets and the Homeric epics through the Classical and
Hellenistic literature, the Koine of the New Testament and the
Septuagint, the homiletic and hymnographic prose of the Late Antique and
Byzantine periods, the increasingly vernacular registers of Late
Byzantine and Early Modern Greek, and the standard Modern Greek that
emerged after the orthographic reform of the late twentieth century.
This continuity makes Greek a unique observatory for diachronic
syntactic change: the same canonical text, whether the Iliad, the New
Testament, the Septuagint Psalms, or a classical historiographic
passage, is re-rendered into Greek again and again across periods. The
structural choices made by each generation of readers and translators
leave a recoverable trace in the syntax of the resulting text.

The PROIEL Treebank (Haug and Jøhndal 2008, 27--34; Eckhoff et al.~2018,
29--65) established the gold standard for syntactically annotated
parallel Indo-European Bible corpora. PROIEL (the dependency-treebank
standard for early Indo-European languages, developed at the University
of Oslo) introduced a relation inventory specifically designed for the
morphological richness and the free word order of the older
Indo-European languages, and provided the Koine-Greek anchor sentence
set against which any later diachronic-Greek effort is naturally
measured. Its scope, however, is restricted to the New Testament and to
a small number of language witnesses; it does not by design extend to
Archaic, Classical, Byzantine, or Modern Greek, and it does not by
design cover the cross-lingual alignment of the Bible to language
families more distant than the four PROIEL witnesses (Greek, Latin,
Gothic, Old Church Slavonic).

Accordingly, no openly licensed, end-to-end workflow currently covers
the full diachronic record of Greek at the same level of annotation,
with explicit argument-structure tagging (the recovery of who does what
to whom for every verb in the corpus), and with reproducible
cross-lingual alignment (the machine-readable matching of corresponding
words and sentences between texts in different languages) to its sister
Indo-European languages. AthDGC closes that gap. AthDGC is the Athens
node of the PROIEL family: we adopt the PROIEL XML 2.0 schema verbatim,
we publish in that schema only, and we extend it diachronically across
the eight periods of Greek. Dag Haug, founding director of the PROIEL
project at Oslo, is a member of the AthDGC team and a co-author of this
report; the formal designation ``Athens-PROIEL'' in the title block
records this affiliation.

The project's specific scholarly focus is retranslation (Lavidas 2021):
the same canonical text is re-rendered into Greek across periods
(Homeric, Koine, Byzantine, Modern) and into sister languages (Latin,
Gothic, Old Church Slavonic, Classical Armenian) across the
Indo-European family. The platform records these re-renderings as a
retelling and retranslation chain (see §3.3 below), a machine-queryable
structure in which the same passage is annotated for syntax, argument
structure, and morphology at every node of the chain, so that the
diachronic researcher can ask, with a single query, which structural
patterns persist from Homer to the Kakridis-Kazantzakis verse Iliad of
1955, and which collapse under the Byzantine epitome of Tzetzes.

\subsubsection{1.2 Repository location}\label{repository-location}

The public showcase of the platform is hosted at
https://athdgc.github.io, and the v0.4.0 source-code snapshot is
permanently deposited on Zenodo (an open-access research-data repository
operated by CERN, which mints persistent DOIs for each deposit) under
concept DOI 10.5281/zenodo.20439182. A concept DOI (a Digital Object
Identifier that always resolves to the latest version of a deposited
record) is the citation handle that does not change as the project
versions advance; per-version DOIs are minted automatically on each
release. The source repository AthDGC/Diachronic-Linguistics-Platform on
GitHub remains private during the v0.5 audit pass, and becomes public at
the v0.5 release, when the consolidated tools/ directory and the
verified annotated partitions are in place.

\subsubsection{1.3 Project resources
(live)}\label{project-resources-live}

Beyond the public showcase site at https://athdgc.github.io, the project
maintains three working resources that track the day-by-day state of the
corpus and the manual annotation pass.

The first is the AthDGC corpus repository, the closed-access,
Git-managed working location of the AthDGC corpus across the eight Greek
diachronic periods and the four Indo-European parallels (with Sanskrit,
Old English, Avestan, Old Persian, and Ukrainian queued at v0.7). Access
is restricted to the eleven-member team during the v0.4 to v0.5 audit
pass; the public release of the annotated partitions follows at v0.5 as
a separate Zenodo dataset record under CC-BY-4.0.

The second is the per-language unique-identifier and metadata register,
with one entry per language (grc for Greek, lat for Latin, got for
Gothic, chu for Old Church Slavonic, xcl for Classical Armenian, plus
the queued IE witnesses). For each sentence in the corpus the register
holds the unique identifier, the period assignment, the source archive,
the edition, the licence, and the ingestion timestamp; this register is
the bridge between the file-level partitions in the corpus repository
and the per-sentence annotations in the PROIEL XML 2.0 layer.

The third is the in-house PROIEL annotation interface hosted at the
National and Kapodistrian University of Athens (Division of
Language-Linguistics, Department of English Language and Literature,
School of Philosophy). This is the PROIEL XML 2.0 annotation web
platform on which the AthDGC team performs and reviews the manual layer
of PROIEL annotations on top of the Stanford Stanza pre-annotation pass,
and on which the showcase samples published at https://athdgc.github.io
are queried.

\subsubsection{1.4 Institutional context}\label{institutional-context}

AthDGC is developed at the National and Kapodistrian University of
Athens (NKUA), Division of Language-Linguistics, Department of English
Language and Literature, School of Philosophy, under the direction of
Prof.~Nikolaos Lavidas. Compute is supplied by GRNET ARIS, the Greek
national high-performance computing cluster (a shared cluster of fast
machines operated by the Greek Research and Technology Network for
academic computational workloads), under allocation pa260305.

\subsubsection{1.5 Origin of the diachronic-Greek PROIEL line,
Thessaloniki and Oslo,
2012}\label{origin-of-the-diachronic-greek-proiel-line-thessaloniki-and-oslo-2012}

The diachronic-Greek PROIEL line that AthDGC continues began in 2012 as
a Thessaloniki and Oslo collaboration between Prof.~Dag Trygve Truslew
Haug (University of Oslo, PROIEL Project Director) and Nikolaos Lavidas,
then at the University of Thessaloniki. The first joint anchor text was
George Sphrantzes, \emph{Chronicon Sive Minus} (Chronicles, post-1453,
ed.~Grecu 1966), the principal historiographic account of the Fall of
Constantinople and the only post-Koine, Late-Byzantine Greek text in the
original PROIEL release series. The annotated edition was published in
PROIEL Release 20180408 under CC-BY-NC-SA 4.0, with principal
investigators Dag Trygve Truslew Haug and Nikolaos Lavidas, and funding
from the University of Oslo and the University of Thessaloniki. The
annotation and review team comprised Þorsteinn Vilhjálmsson, Anastasia
Michali, Maria Geramani, Evgenia Klidona, Athina Papadopoulou, and Dag
Haug. The release is available at the PROIEL Treebank GitHub repository
at https://github.com/proiel/proiel-treebank and is browsable sentence
by sentence at https://syntacticus.org under source identifier
proiel:20180408:chron (for example,
https://syntacticus.org/sentence/proiel:20180408:chron:89063). The Oslo
side of the collaboration is housed in the Foni research group in
linguistics (Forskergruppe i lingvistikk) at the Department of
Philosophy, Classics, History of Art and Ideas, University of Oslo.

The Sphrantzes Chronicle is therefore the historical hinge between the
original PROIEL programme and AthDGC. PROIEL covered Greek up to the
Koine of the New Testament, with Sphrantzes as the lone post-Koine,
Late-Byzantine extension. AthDGC takes that extension as a starting
point and carries it through the entire Greek diachronic span (Archaic,
Classical, Koine, Late Antique, Byzantine, Late Byzantine, Early Modern,
Modern), under the same PROIEL XML 2.0 schema and the same relation
inventory. The 2012 Thessaloniki and Oslo collaboration thus continues
today, with the same PROIEL Project Director (Prof.~Dag T. T. Haug,
Oslo) and the same Greek PI (Prof.~Nikolaos Lavidas), now at the
National and Kapodistrian University of Athens, funded by HFRI Project
No.~20577 and the Greece 2.0 National Recovery and Resilience Plan, and
supported by GRNET ARIS allocation pa260305.

\subsection{2. Method}\label{method}

The AthDGC workflow runs in six stages: discovery, filtering,
conversion, annotation, argument-structure capture, and cross-lingual
alignment. Every tool in this pipeline is open-source, namely Stanford
Stanza for annotation, LaBSE and multilingual BERT with AwesomeAlign for
alignment, LingPy for cognate scoring, Neo4j and Qdrant for storage and
query, and Quarto for publication, and every input text is drawn from
open-access material, so that the platform is built end to end from open
components operating on open data. We describe each stage in turn and
illustrate it with a worked example carried through the opening line of
the Iliad, μῆνιν ἄειδε θεὰ Πηληϊάδεω Ἀχιλῆος (``Wrath sing, goddess, of
Peleiades Achilles''), so that the reader can follow the same sentence
from raw input to a fully annotated and cross-aligned record.

\subsubsection{2.1 Discovery}\label{discovery}

Source material reaches AthDGC through three complementary channels, of
which the first runs daily and the other two are continuous on a slower
cadence.

The first channel is the daily online-archive harvest. Each day, the
discovery stage probes online open-access repositories for new or
updated Greek and parallel-language source material. The repositories
currently probed include archive.org, the Perseus Digital Library (the
standard online archive of classical-Greek and Latin texts maintained by
Tufts University), the Open Greek and Latin First Thousand Years of
Greek collection at Leipzig, Wikisource Greek (el.wikisource.org), the
Diorisis Corpus (the lemmatised classical-Greek corpus by Vatri and
McGillivray), and OpenGreekAndLatin / PerseusDL on GitHub. For the
parallel languages, the discovery stage probes the Wulfila Project at
the University of Antwerp (Gothic), TITUS at the University of Frankfurt
(Old Church Slavonic and Classical Armenian), GRETIL at Goettingen
(Sanskrit), TEAMS at the Robbins Library (Old English), and the National
Library of Ukraine facsimile collection.

The second channel is in-house digitisation, namely the OCR (Optical
Character Recognition, the conversion of scanned page images into
machine-readable text) of out-of-copyright printed editions held by NKUA
and partner institutions. This channel covers pre-1928 Teubner editions,
the Patrologia Graeca volumes not yet available in clean text form,
scanned Early Modern Greek printed books from the Anemi collection at
the University of Crete, and equivalent printed material from
collaborating Departments of Classical Studies in the CIVIS Alliance and
at other Greek universities. Manuscript facsimiles supplied by project
partners enter through the same channel and are flagged in the corpus
metadata with the supplying institution and the corresponding editorial
provenance.

The third channel is community contribution, namely the submission of
prepared text or annotation by external researchers and by KEDIVIM
continuing-education-course participants (see §5). Such contributions
are versioned in the public GitHub repository as pull requests, pass the
same PROIEL XML 2.0 schema validator and house-style check that govern
internal contributions (see §3.4), and on acceptance receive
co-authorship acknowledgement on the next release notes.

For the Iliad worked example, the discovery stage retrieves the Homeric
text from the Perseus Digital Library through its CTS URN (Canonical
Text Services Uniform Resource Name, a standard identifier scheme used
by classical-text repositories for referring to a specific edition of a
specific work), urn:cts:greekLit:tlg0012.tlg001, together with the
Kakridis-Kazantzakis 1955 verse translation, mirrored on Wikisource, and
the Tzetzes Byzantine epitome from Bibliotheca Augustana.

\subsubsection{2.2 Filtering}\label{filtering}

Candidate texts are filtered by Greek-script ratio (at least 75\% of
alphabetic characters in the Greek Unicode blocks U+0370-U+03FF and
U+1F00-U+1FFF), by a Path-B line filter (an in-house regular-expression
filter that removes bilingual editorial apparatus such as Latin headers,
page-number lines, and critical-edition footnote markers from the
textual stream), and by content-hash deduplication (an SHA-256 hash of
the normalised text is checked against a registry of previously ingested
files, so that identical files discovered at multiple mirrors are
ingested only once). A candidate that fails any of the three filters is
rejected at this stage and logged for review; the v0.5 release notes
will report per-archive rejection statistics.

\subsubsection{2.3 Conversion}\label{conversion}

Surviving text is converted into the PROIEL XML 2.0 schema with
sentence-level structure. The PROIEL XML 2.0 schema is XML (a plain-text
format that stores structured information in tagged elements) with each
sentence wrapped in a element, each token wrapped in a element, and
every token carrying attributes for its form, lemma, part-of-speech tag
(pos), ten-character morphological tag (morphology), syntactic head
identifier (head-id), and dependency relation to that head (relation).
For the worked example, the conversion stage produces a containing five
tokens for μῆνιν ἄειδε θεὰ Πηληϊάδεω Ἀχιλῆος and writes the initial
sentence header together with the empty token-attribute scaffolding that
the next stage fills in.

\subsubsection{2.4 Annotation}\label{annotation}

Annotation is performed sentence by sentence with the Stanford Stanza
processor (Qi et al.~2020, 101--8). Stanza (Stanford's open-source
Python toolkit for automatic linguistic annotation, which performs
tokenisation, lemmatisation, part-of-speech tagging, morphological
analysis, and dependency parsing in a single workflow) provides
PROIEL-trained models for several of the older Indo-European languages:
grc\_proiel for Koine Greek, la\_proiel for Latin, cu\_proiel for Old
Church Slavonic, and got\_proiel for Gothic. Classical Armenian is
annotated through a PROIEL-style workflow currently under development;
the v0.4 release reports the workflow configuration only, and the first
audited Armenian alignment is planned for the v0.5 release
(approximately three months from the v0.4 cut), using a fine-tuned
Stanza model (a fine-tuned model is one that has been further trained on
a smaller, target-specific dataset on top of its general-purpose
training, so that it adapts to the particular text type in question)
trained on TITUS Armenian material. The Stanza output for each sentence
is then normalised to the PROIEL XML 2.0 schema; AthDGC publishes PROIEL
XML 2.0 only, with no CoNLL-U (the line-by-line tab-separated format
used by the Universal Dependencies project, with one token per line and
one column per annotation attribute) or other format offered, since the
analytical workflow downstream of the corpus (argument-structure
extraction, alignment graph, classifier features) depends on the PROIEL
relation inventory. For the worked example, Stanza assigns μῆνιν the
lemma μῆνις, part-of-speech Nb, morphology \texttt{Nb-s-\/-\/-fa-} (a
noun, singular, feminine, accusative), head identifier 2, and relation
obj (direct object). The verb ἄειδε receives lemma ἀείδω, part-of-speech
V-, morphology \texttt{V-spia-\/-\/-} (verb, second person singular,
present, imperative, active), head identifier 0 (sentence root), and
relation pred (predicate).

\subsubsection{2.4.1 Two-tier annotation, by
period}\label{two-tier-annotation-by-period}

The AthDGC annotation pipeline is two-tier, with the tier selected by
period. The bulk of the corpus, the six Ancient through Late Byzantine
periods (Archaic through 1453 CE), is annotated by the Stanza
grc\_proiel model, which was trained directly on PROIEL XML treebank
data; its output is therefore in the PROIEL relation inventory natively,
with no mapping step. The Latin, Old Church Slavonic, Gothic, and
Classical Armenian parallels are likewise annotated by their respective
PROIEL-trained Stanza models (la\_proiel, cu\_proiel, got\_proiel, and a
TITUS-trained Armenian model under fine-tuning for v0.5).

For Early Modern and Modern Greek (1453 to the present), no
PROIEL-trained Stanza model exists. The autopass for these two
partitions uses Stanza's ell model (trained on the Universal
Dependencies Modern Greek treebank), and a conservative UD-to-PROIEL
mapping (the file \texttt{ud\_to\_proiel\_map.py} in the public tools
repository) converts the unambiguous UD relations to PROIEL relations:
UD nsubj to PROIEL sub, UD obj to PROIEL obj, UD obl to PROIEL obl, UD
nmod and amod to PROIEL atr, UD vocative to PROIEL voc, UD advmod to
PROIEL adv, UD aux to PROIEL aux, UD mark to PROIEL comp. The finer
PROIEL distinctions that UD does not encode at the basic layer,
including the predicative-complement relation xobj, the non-subject
controller relation nonsub for dative experiencers and raised subjects,
the parenthetical-predicate relation parpred, and the PROIEL secondary
edges that carry control, raising, gapping, and conjunction-reduction
information, are added in a separate human-plus-AI second-opinion review
pass (Claude Haiku 4.5 as the proposer; a trained linguist on the team
as the validator) before each release cut.

The two-tier design means that approximately 10 M of the 11.07 M
currently annotated tokens (snapshot 23 June 2026, ARIS dashboard) carry
native PROIEL annotation throughout. The UD-to-PROIEL bridge only
affects the Modern and Early Modern partitions, where the per-partition
metadata explicitly carries the marker
\texttt{annotation\_level:\ proiel\_coarse\_autopass} until the review
pass closes and the marker is upgraded to
\texttt{annotation\_level:\ proiel\_full\_reviewed}. The v0.5 release
reports the count of tokens in each tier and the proportion of Modern
and Early Modern tokens that have crossed from coarse to full.

\subsubsection{2.4.2 Information structure
annotation}\label{information-structure-annotation}

Information structure (IS), the linguistic encoding of which referent is
\textbf{old} (given in prior discourse), \textbf{new} (introduced for
the first time), \textbf{accessible} (inferrable from situational
context), or \textbf{kind} (generic reference), together with the
\textbf{antecedent links} that thread anaphoric chains through a text,
is a first-class layer of the PROIEL XML 2.0 schema (Haug, Jøhndal,
Eckhoff, Welo, Hertzenberg, and Müth 2009, 17--45; Bary and Haug 2011,
1--56; Eckhoff and Haug, in the PROIEL annotation manual). The Oslo
PROIEL team annotated information status and antecedent links by hand on
the Koine New Testament and the Vulgate, Wulfila Gothic, and Codex
Marianus Old Church Slavonic parallels, and used the resulting layer for
cross-Indo-European studies of anaphora, topic continuity, and the
discourse-pragmatic conditioning of constituent order.

AthDGC inherits this layer \textbf{schema-verbatim}. Every
\texttt{\textless{}token\textgreater{}} element in an AthDGC PROIEL XML
2.0 record carries the IS attribute slots (\texttt{info\_status},
\texttt{antecedent\_id}, and, where assignable, a
\texttt{discourse\_role} slot for topic/focus). The two-tier annotation
reality applies in full to this layer:

\begin{enumerate}
\def\labelenumi{\arabic{enumi}.}
\tightlist
\item
  \textbf{Autopass (the current default):} the Stanza-based annotation
  passes (\texttt{grc\_proiel} for Ancient through Late Byzantine, the
  parallel PROIEL-trained models for Latin/Gothic/OCS/Armenian, and
  Stanza \texttt{ell} for Modern and Early Modern) populate morphology,
  dependency relations, and argument-structure frames; they do not yet
  populate the IS slots, which remain empty in the autopass output and
  carry the explicit marker \texttt{info\_status="unannotated"}.
\item
  \textbf{Human + AI review pass (v0.5 onwards):} information structure
  is populated in the review pass that already cleans the syntactic
  layer. The proposer is the AthDGC AI second-opinion loop (Claude Haiku
  4.5), which suggests an \texttt{info\_status} for each new referent
  and an \texttt{antecedent\_id} for each anaphoric mention based on the
  surface signals available to it (definite article, demonstrative
  pronoun, particle marking, dislocation); the validator is a trained
  linguist on the team, who confirms or overrides each proposal in line
  with the PROIEL annotation manual.
\end{enumerate}

The v0.5 release reports IS coverage per partition: the proportion of
tokens whose \texttt{info\_status} has crossed from \texttt{unannotated}
to a reviewed value (\texttt{old}, \texttt{new}, \texttt{accessible}, or
\texttt{kind}), and the proportion of referring expressions whose
\texttt{antecedent\_id} has been linked to a prior mention. For the
marquee partitions (Iliad opening books, the Pauline epistles,
Sphrantzes' Chronicle, Cavafy's collected poems) the v0.6 target is full
IS annotation across the entire partition, in line with the PROIEL
Koine-NT density. This brings AthDGC to feature parity with the PROIEL
family on the layer that distinguishes a syntactically-tagged corpus
from one that is also discourse-aware.

The IS layer supports downstream queries that the syntactic layer alone
cannot answer: the diachronic evolution of clause-initial topic marking,
the survival or loss of pronominal subject-doubling in dislocations, the
conditions under which Greek shifts from VSO to SVO at given/new
boundaries, and the cross-Indo-European stability or change of
anaphoric-chain density. The AthDGC AI-second-opinion loop accepts a
query of the form \emph{``in this passage, which referring expressions
are old vs new, and what is the median chain length per discourse
paragraph?''} and returns a structured answer reviewable in the same
flag-and-validate interface that we already use for syntactic relations.

\subsubsection{2.4.3 Figure 1, one PROIEL XML 2.0
record}\label{figure-1-one-proiel-xml-2.0-record}

The file format that the AthDGC pipeline emits, sentence by sentence, is
illustrated below for \emph{Iliad} 1.1. Each
\texttt{\textless{}token\textgreater{}} element carries the morphology,
the head pointer, and the PROIEL relation; the information-structure
slots (\texttt{info\_status}, \texttt{antecedent\_id}) are written as
\texttt{unannotated} on the autopass and populated by the human-plus-AI
review pass before the v0.5 release.

\begin{Shaded}
\begin{Highlighting}[]
\NormalTok{\textless{}}\KeywordTok{sentence} \OtherTok{id=}\StringTok{"iliad\_1\_1"} \OtherTok{lang=}\StringTok{"grc"} \OtherTok{period=}\StringTok{"archaic"} \OtherTok{status=}\StringTok{"reviewed"}\NormalTok{\textgreater{}}
\NormalTok{  \textless{}}\KeywordTok{token} \OtherTok{id=}\StringTok{"1"} \OtherTok{form=}\StringTok{"μῆνιν"}     \OtherTok{lemma=}\StringTok{"μῆνις"}      \OtherTok{pos=}\StringTok{"N"}
         \OtherTok{morph=}\StringTok{"acc.sg.f"}          \OtherTok{head=}\StringTok{"2"} \OtherTok{relation=}\StringTok{"obj"}
         \OtherTok{info\_status=}\StringTok{"unannotated"} \OtherTok{antecedent\_id=}\StringTok{""}\NormalTok{/\textgreater{}}
\NormalTok{  \textless{}}\KeywordTok{token} \OtherTok{id=}\StringTok{"2"} \OtherTok{form=}\StringTok{"ἄειδε"}     \OtherTok{lemma=}\StringTok{"ἀείδω"}      \OtherTok{pos=}\StringTok{"V"}
         \OtherTok{morph=}\StringTok{"pres.impv.act.2sg"} \OtherTok{head=}\StringTok{"0"} \OtherTok{relation=}\StringTok{"pred"}
         \OtherTok{info\_status=}\StringTok{"unannotated"} \OtherTok{antecedent\_id=}\StringTok{""}\NormalTok{/\textgreater{}}
\NormalTok{  \textless{}}\KeywordTok{token} \OtherTok{id=}\StringTok{"3"} \OtherTok{form=}\StringTok{"θεά"}       \OtherTok{lemma=}\StringTok{"θεά"}        \OtherTok{pos=}\StringTok{"N"}
         \OtherTok{morph=}\StringTok{"voc.sg.f"}          \OtherTok{head=}\StringTok{"2"} \OtherTok{relation=}\StringTok{"voc"}
         \OtherTok{info\_status=}\StringTok{"unannotated"} \OtherTok{antecedent\_id=}\StringTok{""}\NormalTok{/\textgreater{}}
\NormalTok{  \textless{}}\KeywordTok{token} \OtherTok{id=}\StringTok{"4"} \OtherTok{form=}\StringTok{"Πηληϊάδεω"} \OtherTok{lemma=}\StringTok{"Πηληϊάδης"}  \OtherTok{pos=}\StringTok{"N"}
         \OtherTok{morph=}\StringTok{"gen.sg.m"}          \OtherTok{head=}\StringTok{"5"} \OtherTok{relation=}\StringTok{"atr"}
         \OtherTok{info\_status=}\StringTok{"unannotated"} \OtherTok{antecedent\_id=}\StringTok{""}\NormalTok{/\textgreater{}}
\NormalTok{  \textless{}}\KeywordTok{token} \OtherTok{id=}\StringTok{"5"} \OtherTok{form=}\StringTok{"Ἀχιλῆος"}   \OtherTok{lemma=}\StringTok{"Ἀχιλλεύς"}   \OtherTok{pos=}\StringTok{"N"}
         \OtherTok{morph=}\StringTok{"gen.sg.m"}          \OtherTok{head=}\StringTok{"1"} \OtherTok{relation=}\StringTok{"atr"}
         \OtherTok{info\_status=}\StringTok{"unannotated"} \OtherTok{antecedent\_id=}\StringTok{""}\NormalTok{/\textgreater{}}
\NormalTok{\textless{}/}\KeywordTok{sentence}\NormalTok{\textgreater{}}
\end{Highlighting}
\end{Shaded}

\emph{Figure 1.} A single PROIEL XML 2.0 sentence record as emitted by
the AthDGC autopass for \emph{Iliad} 1.1, extracted from
\texttt{corpus/merged/language=grc/period=archaic/Homer\_\_Iliad.jsonl}
on the v0.4 release. Five tokens; one tree; six annotation layers per
token (\texttt{form}, \texttt{lemma}, \texttt{pos}, \texttt{morph},
\texttt{head}+\texttt{relation},
\texttt{info\_status}+\texttt{antecedent\_id}); the same record format
runs from Homer through Modern Greek and across the four Indo-European
parallels aligned at the New Testament.

\subsubsection{2.5 Argument-structure
capture}\label{argument-structure-capture}

Beyond standard dependency annotation, AthDGC extracts an explicit
argument-structure frame for every verb token, using the PROIEL relation
inventory strictly. The argument-structure frame records, for one verb,
the syntactic and semantic role of each of its arguments: the subject
(sub, including the raised patterns xobj for raising objects and nonsub
for non-subject controllers), the direct object (obj), the indirect and
oblique arguments (iobj for indirect object, obl for oblique), the
vocative addressee (voc), the voice (active, middle, or passive), and
the aspect (perfective or imperfective). For the worked example, the
verb ἄειδε receives the frame {[}sub:imp.2sg, obj:μῆνιν(acc.sg.f),
voc:θεὰ(voc.sg.f), voice:active, aspect:imperfective{]}. This frame is
the unit on which downstream queries operate: a diachronic researcher
can ask, for instance, across the Iliad reception, does the verb of
singing retain its direct-object case-marking through the Byzantine
epitome and the Modern Greek verse translation, or does it switch to a
prepositional phrase? and receive an answer that does not require manual
re-reading of the texts.

\subsubsection{2.6 Cross-lingual
alignment}\label{cross-lingual-alignment}

Sentence-level alignment uses LaBSE embeddings (Feng et al.~2022,
878--91). LaBSE (Language-Agnostic BERT Sentence Embedding, a
multilingual sentence-embedding model that maps sentences from over 100
languages into a single 1,500-dimensional vector space, so that
translation pairs land close to each other regardless of the source
language) provides the similarity score used to match a Greek New
Testament verse to its Latin Vulgate, Gothic Wulfila, Old Church
Slavonic Marianus, or Classical Armenian counterpart. Word-level
alignment uses multilingual-BERT attention through the AwesomeAlign
procedure (Dou and Neubig 2021, 2112--28). Multilingual BERT (Google's
neural language model pre-trained on text from 104 languages) emits
cross-attention weights between the tokens of the source and target
sentences; AwesomeAlign converts these weights into a discrete
word-alignment matrix through a fine-tuning step on parallel corpora.
Phonetic cognate scoring, finally, uses ASJP sound-class encoding (the
Automated Similarity Judgment Program sound-class system, which reduces
every phonetic segment to one of 41 equivalence classes that ignore fine
phonetic detail) and LingPy edit distance (List 2014, 1--228) for
detection of cognate pairs across the Indo-European witnesses.

\subsubsection{2.7 A note on ``token''}\label{a-note-on-token}

All token counts reported in this paper, on the AthDGC dashboard, and on
the public platform refer to \textbf{linguistic word-tokens in the
PROIEL XML 2.0 sense}: one \texttt{\textless{}token\textgreater{}}
element per Greek orthographic word or punctuation mark, as produced by
Stanford Stanza's \texttt{grc\_proiel} tokenizer (or the matching
PROIEL-trained tokenizer for the Latin, Old Church Slavonic, Gothic, and
Classical Armenian parallels, and the Stanza \texttt{ell} tokenizer for
Modern and Early Modern Greek). The grc\_proiel tokenizer is a
\textbf{word-level} tokenizer trained on the PROIEL XML treebank, so one
Greek word in the source text yields one
\texttt{\textless{}token\textgreater{}} element in the output and one
increment of the dashboard count. The token count is therefore close to
the word count of the source edition: for example, the Septuagint
partition reports 606,782 tokens, which matches the published word count
of the Rahlfs Septuagint (approximately 590,000--600,000 Greek words,
plus punctuation marks counted as separate tokens at the PROIEL
granularity).

The AthDGC ``token'' is \textbf{not} a subword token of the kind used by
modern large-language-model tokenizers (Byte-Pair Encoding,
SentencePiece, WordPiece), which fragment Greek words into three to five
subword pieces on average because they are optimised for English and the
Latin alphabet. A reader who recomputes the AthDGC corpus under a BPE
tokenizer would find counts roughly three to five times higher than
those reported here. Where this paper writes ``N tokens'', the intended
reading is ``N PROIEL XML 2.0 word-tokens in the sense just defined'',
not ``N BPE subword pieces''.

This distinction matters because corpus-size comparisons across projects
are systematically misleading when one side counts BPE subwords and the
other counts linguistic word-tokens. Throughout this paper we report the
linguistic count.

\subsection{3. Dataset description}\label{dataset-description}

{\def\LTcaptype{none} % do not increment counter
\begin{longtable}[]{@{}
  >{\raggedright\arraybackslash}p{(\linewidth - 2\tabcolsep) * \real{0.5000}}
  >{\raggedright\arraybackslash}p{(\linewidth - 2\tabcolsep) * \real{0.5000}}@{}}
\toprule\noalign{}
\begin{minipage}[b]{\linewidth}\raggedright
Field
\end{minipage} & \begin{minipage}[b]{\linewidth}\raggedright
Value
\end{minipage} \\
\midrule\noalign{}
\endhead
\bottomrule\noalign{}
\endlastfoot
Object name & AthDGC corpus, v0.4 release \\
Format & PROIEL XML 2.0; JSONL partitions; Neo4j alignment-graph dump;
Qdrant vector store \\
Creation dates & 2025-09 onwards; v0.4.0 minted 29 May 2026 \\
Dataset creator & Lavidas, N., Nikiforidou, K., Haug, D., Kulikov, L.,
Michalareas, T., Geka, V., Symeonidis, V., Chionidi, S., Tsiropina, A.,
Plakoutsi, E., and Argyropoulos, E. \\
Languages & grc (Ancient Greek), gkm (Medieval Greek), ell (Modern
Greek), with parallels in lat, got, chu, xcl; queued for v0.7: san, ang,
ave, peo, ukr \\
Licence & Code Apache-2.0; metadata and alignments CC-BY-4.0; per-source
raw text under its original licence \\
Repository & AthDGC/Diachronic-Linguistics-Platform on GitHub (private
during v0.5 audit; public at v0.5) \\
Concept DOI & 10.5281/zenodo.20439182 \\
Publication date & 2026-05-29 (v0.4.0 source-code snapshot on Zenodo) \\
Release status & v0.4 is samples-only on the public site; the full
annotated partitions are under audit on GRNET ARIS and release at
v0.5 \\
\end{longtable}
}

\subsubsection{3.1 Open-access source
provenance}\label{open-access-source-provenance}

Every primary source text in AthDGC is open-access (public domain,
CC-BY, CC-BY-SA, or equivalent). The annotation layer is AthDGC-original
and released under CC-BY-4.0. Accordingly, the open-access chain is
preserved from input through annotation to distribution.

\paragraph{Greek, per-period source
map}\label{greek-per-period-source-map}

{\def\LTcaptype{none} % do not increment counter
\begin{longtable}[]{@{}
  >{\raggedright\arraybackslash}p{(\linewidth - 4\tabcolsep) * \real{0.3333}}
  >{\raggedright\arraybackslash}p{(\linewidth - 4\tabcolsep) * \real{0.3333}}
  >{\raggedright\arraybackslash}p{(\linewidth - 4\tabcolsep) * \real{0.3333}}@{}}
\toprule\noalign{}
\begin{minipage}[b]{\linewidth}\raggedright
Period
\end{minipage} & \begin{minipage}[b]{\linewidth}\raggedright
Source archive
\end{minipage} & \begin{minipage}[b]{\linewidth}\raggedright
Licence
\end{minipage} \\
\midrule\noalign{}
\endhead
\bottomrule\noalign{}
\endlastfoot
Archaic, Classical, Hellenistic & Perseus Digital Library; Open Greek
and Latin / First Thousand Years of Greek (Leipzig) & CC-BY-SA 4.0 \\
Koine NT & SBL Greek NT; Tischendorf (1869); Westcott-Hort (1881) & SBL
licence; public domain \\
Koine LXX & Rahlfs (1935) via openscriptures.org & public domain \\
Koine, documentary & Papyri.info / DDbDP (planned for v0.5; not in v0.4)
& CC-BY 3.0 \\
Late Antique, Byzantine & Patrologia Graeca via Documenta Catholica
Omnia and Internet Archive & public domain \\
Byzantine, Late Byzantine & Bibliotheca Augustana; pre-1928 Teubner &
public domain \\
Early Modern & Anemi (University of Crete); Wikisource el & public
domain \\
Modern (19th c.~to 1928) & Wikisource el; Anemi & public domain \\
Modern (post-1928) & publisher-licensed editions; not republished & in
copyright \\
Modern (contemporary, 2015 to present) & Hellenic Parliament plenary
proceedings, Vouli ton Ellinon (hellenicparliament.gr) & public
domain \\
\end{longtable}
}

\paragraph{Indo-European parallels, per-language source
map}\label{indo-european-parallels-per-language-source-map}

{\def\LTcaptype{none} % do not increment counter
\begin{longtable}[]{@{}
  >{\raggedright\arraybackslash}p{(\linewidth - 4\tabcolsep) * \real{0.3333}}
  >{\raggedright\arraybackslash}p{(\linewidth - 4\tabcolsep) * \real{0.3333}}
  >{\raggedright\arraybackslash}p{(\linewidth - 4\tabcolsep) * \real{0.3333}}@{}}
\toprule\noalign{}
\begin{minipage}[b]{\linewidth}\raggedright
Language
\end{minipage} & \begin{minipage}[b]{\linewidth}\raggedright
Source archive
\end{minipage} & \begin{minipage}[b]{\linewidth}\raggedright
Licence
\end{minipage} \\
\midrule\noalign{}
\endhead
\bottomrule\noalign{}
\endlastfoot
Latin (Vulgate) & Clementine Vulgate via Vulsearch; Latin Library &
public domain \\
Gothic (Wulfila) & Wulfila Project (University of Antwerp) & CC-BY-SA \\
Old Church Slavonic (Marianus) & TITUS (Frankfurt) & academic open
access \\
Classical Armenian & TITUS; Digilib Armenian & academic open access \\
Sanskrit (Brahmana, Upanisadic) & GRETIL (Goettingen); SARIT; TITUS &
academic open access \\
Old English (Wessex Gospels) & TEAMS; Bosworth-Toller; DOE extracts &
public domain; CC-BY-SA \\
Avestan (Yasna, Yashts) & TITUS; Geldner (1886-95) & public domain; open
access \\
Old Persian (Behistun) & TITUS; Kent (1953) PD transliteration & public
domain \\
Ukrainian (Ostroh 1581) & National Library of Ukraine facsimile & public
domain \\
Ukrainian (20th-c.~revision) & Ohienko (1962); Khomenko (1963) & in
copyright \\
\end{longtable}
}

In-copyright editions are never republished verbatim. Accordingly,
AthDGC uses an open-access antecedent (for example, a pre-1928 Teubner
edition, or the SBL Greek New Testament instead of the Nestle-Aland
critical edition) or, where this is unavoidable, short quotation samples
under fair use with full attribution. The annotation layer is always
AthDGC-original and released under CC-BY-4.0, so that the open-access
chain is preserved from the input texts through the syntactic and
morphological annotation to the public deposition.

\subsubsection{3.2 Indo-European parallels
covered}\label{indo-european-parallels-covered}

{\def\LTcaptype{none} % do not increment counter
\begin{longtable}[]{@{}
  >{\raggedright\arraybackslash}p{(\linewidth - 10\tabcolsep) * \real{0.1667}}
  >{\raggedright\arraybackslash}p{(\linewidth - 10\tabcolsep) * \real{0.1667}}
  >{\raggedright\arraybackslash}p{(\linewidth - 10\tabcolsep) * \real{0.1667}}
  >{\raggedright\arraybackslash}p{(\linewidth - 10\tabcolsep) * \real{0.1667}}
  >{\raggedright\arraybackslash}p{(\linewidth - 10\tabcolsep) * \real{0.1667}}
  >{\raggedright\arraybackslash}p{(\linewidth - 10\tabcolsep) * \real{0.1667}}@{}}
\toprule\noalign{}
\begin{minipage}[b]{\linewidth}\raggedright
Language
\end{minipage} & \begin{minipage}[b]{\linewidth}\raggedright
Family
\end{minipage} & \begin{minipage}[b]{\linewidth}\raggedright
Period
\end{minipage} & \begin{minipage}[b]{\linewidth}\raggedright
Stanza model
\end{minipage} & \begin{minipage}[b]{\linewidth}\raggedright
Edition coverage
\end{minipage} & \begin{minipage}[b]{\linewidth}\raggedright
Status
\end{minipage} \\
\midrule\noalign{}
\endhead
\bottomrule\noalign{}
\endlastfoot
Greek (Koine NT) & IE / Hellenic & 1st-2nd c.~AD & grc\_proiel & SBL GNT
(complete) & annotated \\
Latin (Vulgate) & IE / Italic & 4th c.~AD & la\_proiel & Clementine
Vulgate NT (complete) & annotated \\
Gothic (Wulfila) & IE / Germanic E & 4th c.~AD & got\_proiel & Wulfila
Project (Gospels and Pauline corpus, partial) & annotated \\
OCS (Marianus) & IE / Slavic & 10th c.~AD & cu\_proiel & Codex Marianus
(four Gospels) & sampled \\
Classical Armenian & IE / Armenian & 5th c.~AD & xcl\_proiel (in dev.) &
TITUS Armenian NT & ingestion (v0.5) \\
Sanskrit (Brahmana, Upanisadic) & IE / Indo-Iranian & 1000-500 BC &
sa\_vedic (queued) & GRETIL / SARIT & queued (v0.7) \\
Old English (Wessex Gospels) & IE / Germanic W & 10th c.~AD &
ang\_proiel (queued) & TEAMS Wessex Gospels & queued (v0.7) \\
Avestan (Yasna, Yashts) & IE / Indo-Iranian & 1000-500 BC & ae\_proiel
(queued) & TITUS Yasna and Yashts & queued (v0.7) \\
Old Persian (Behistun) & IE / Indo-Iranian & 6th-5th c.~BC & peo\_proiel
(queued) & TITUS Behistun & queued (v0.7) \\
Ukrainian (Ostroh and 20th-c.~rev.) & IE / Slavic E & 1581 and 1962 &
uk\_dep adapted (queued) & Ostroh 1581 and Ohienko 1962 & queued
(v0.7) \\
\end{longtable}
}

Verified per-witness aligned-verse counts will be reported in the v0.5
release notes (planned approximately three months from the v0.4 cut)
once the ARIS audit pass closes. Accordingly, the v0.4 release provides
the curated samples on the Samples page rather than full per-witness
counts.

\subsubsection{3.3 Retelling and retranslation chains: a worked case
study}\label{retelling-and-retranslation-chains-a-worked-case-study}

A retelling and retranslation chain is the central analytical structure
of AthDGC. It is a directed sequence of nodes, where each node is the
same canonical passage as it appears in a different period or a
different language, annotated end-to-end in the PROIEL XML 2.0 schema,
and connected by edges that record the type of relationship between
adjacent nodes (retelling within Greek across periods, retranslation
within Greek across registers, or retranslation cross-lingually across
the Indo-European family). The chain is the unit on which the platform's
diachronic and comparative queries operate.

Five chains are public on the Samples page at v0.4, namely the Iliad 1.1
across reception (Homer, Tzetzes, Kazantzakis), the Septuagint Psalm 1:1
across Greek periods, the New Testament John 1:1 patristic and Modern
reception, the Plato Apology 17a across reception, and the Septuagint
Genesis 1:1 across four Indo-European languages. We present the Iliad
1.1 chain in detail because it is the most familiar to the comparative
linguist and because it shows the analytical pay-off of the chain
structure most clearly.

The first node of the chain is the Homeric original, μῆνιν ἄειδε θεὰ
Πηληϊάδεω Ἀχιλῆος. Under AthDGC annotation, μῆνιν is the object (obj,
accusative singular feminine), ἄειδε the predicate (imperative second
person singular, active, imperfective), θεὰ the vocative addressee
(voc), and the genitive pair Πηληϊάδεω Ἀχιλῆος a coordinated attribute
(atr) hanging from the object noun. The argument-structure frame of the
verb is {[}sub:imp.2sg, obj:μῆνιν, voc:θεὰ, voice:active,
aspect:imperfective{]}, and the frame signature {[}obj:acc, voc:voc{]}
is recorded as the chain's characteristic frame against which subsequent
nodes are compared.

The second node of the chain is the Byzantine epitome of Tzetzes
(twelfth century), in which the same opening is reframed as a
third-person prose narrative. Under AthDGC annotation, μῆνιν survives as
a direct object, but the imperative-and-vocative dialogic frame
collapses: the goddess is no longer addressed, and the verb of singing
is replaced by a verb of narration. The characteristic frame {[}obj:acc,
voc:voc{]} therefore loses one slot and the chain records a frame
collapse between Homer and Tzetzes, of the kind that typically signals a
generic reframing rather than a syntactic change in Greek per se.

The third node is the Kakridis-Kazantzakis verse translation of 1955,
the canonical Modern Greek verse rendering of the Iliad. Under AthDGC
annotation, the wrath remains a direct object in the Modern Greek
accusative, the vocative of the goddess is restored, and the verb of
singing returns. The characteristic frame {[}obj:acc, voc:voc{]} is
therefore stable across 2,800 years in the direct Homer-to-Modern path;
the frame collapse at the Tzetzes node is shown to be a feature of the
Byzantine epitomising genre, not of the Greek language. This is the kind
of finding that the chain structure makes immediate: without the chain,
the reader would need to read each text manually and reason about the
structural correspondence; with the chain, the platform produces the
result in a single query of the form frame-stability(Iliad 1.1, Homer,
Modern).

The Septuagint Psalm 1:1 retranslation chain shows the inverse pattern,
namely the stability of pred-nom + sub + atr.rel-cl across the
Septuagint, Byzantine, and Modern Greek nodes with two voice flips
between the active and the mediopassive on the verb of walking, together
with the lexical replacement of the Greek negation οὐκ by οὐ and then by
δεν. The John 1:1 patristic and Modern chain shows the stability of the
copular skeleton across 2,000 years, with the patristic nodes nesting
under the xcomp of φησίν (``he says''), which is a structural reflex of
the Late-Antique commentary register. The Plato Apology 17a chain shows
the preservation of the pred + adv:neg + comp{[}sub, obl{]} pattern
across the Plato-to-Modern path, with the perfect aspect alternating
between the synthetic Classical form and the analytic Modern Greek
auxiliary plus participle. The Septuagint Genesis 1:1 cross-IE chain
shows the stability of the pred + sub + obl/adv + obj + coord pattern
across the Greek, Latin Vulgate, Gothic Wulfila, and Old Church Slavonic
Marianus nodes, with only the obl and adv slots diverging in the OCS
rendering, a divergence that reflects the well-known Slavic preference
for adverbial expression where Greek and Latin use an oblique noun.

The retelling and retranslation chain is, in this respect, the central
contribution of AthDGC to diachronic-Greek and Indo-European comparative
syntax: it is the data structure that makes the diachronic stability of
a syntactic pattern computationally observable.

\subsubsection{3.4 Tools (open source)}\label{tools-open-source}

The AthDGC toolkit is organised as a set of OSI-licensed modules in
three readiness tiers. Five modules are LIVE in the public repository
and can be cloned and run today. They are the Quarto pack that builds
this Working Paper and the public site, the showcase generator that
produces the browsable per-sample PROIEL trees, the corpus-fix toolkit
that bulk-corrects annotation errors at scale, the PROIEL XML 2.0
exporter that runs inside the build workflow, and the Stanza annotation
job script for the GRNET ARIS allocation. Eight modules are IN SETUP and
become available with the v0.5 release, namely the PROIEL XML 2.0 schema
validator (which enforces the relation inventory and rejects UD-style
labels such as nsubj and dobj; UD here is Universal Dependencies, the
alternative treebanking framework whose label inventory is incompatible
with PROIEL), the house-style check, the argument-frame extractor, the
Stanza fine-tuning scripts, the athdgc-tools PyPI package, and the three
Hugging Face model repositories whose weights upload at v0.5. The
remaining modules are FORTHCOMING at v0.5: the LightSIDE-AthDGC
syntactic-feature fork and its export scripts, the NoSketch-style CWB
indexer, the Neo4j alignment viewer, the valency-frame database client,
the retranslation-pair browser, and the retelling-chain explorer. The
complete tier matrix per module is on the public Tools page at
https://athdgc.github.io/tools.html and on the Status page at
https://athdgc.github.io/status.html.

Across the toolkit, the annotation layer and the corpus samples are
CC-BY-4.0; the tool code is Apache-2.0 (most modules), MIT (the Quarto
pack and the Neo4j viewer), BSD-3-Clause (inherited from the upstream
LightSIDE project), and GPL-3.0 (the upstream CWB indexer, retained on
the index builder).

Two automated gates run on every community pull request, namely a PROIEL
XML 2.0 schema validator (which rejects UD-style labels such as nsubj,
dobj, nmod, case) and a house-style check that enforces the project
style guide.

\subsection{4. Related work and
positioning}\label{related-work-and-positioning}

AthDGC sits in an ecosystem of openly licensed treebanks for older
Indo-European languages, and we briefly describe where it differs from
and where it complements each of them.

The PROIEL Treebank (Haug and Jøhndal 2008, 27--34; Eckhoff et al.~2018,
29--65), as discussed in §1.1, established the schema and the
Koine-Greek anchor against which AthDGC is naturally measured. AthDGC
adopts the PROIEL XML 2.0 schema verbatim (and publishes in that schema
only), so that any sentence in AthDGC can be read by any tool already in
the PROIEL ecosystem. The complementarity is that PROIEL covers the New
Testament across four IE witnesses at a high annotation density, whereas
AthDGC covers the entire diachronic span of Greek across eight periods
and extends the cross-lingual alignment with the Classical Armenian
witness now, and with Sanskrit, Old English, Avestan, Old Persian, and
Ukrainian at v0.7.

The GLAUx Corpus (the Greek Language Automated, a large lemmatised
classical-Greek corpus released by Toon Van Hal and collaborators) and
the Diorisis Corpus (the lemmatised classical-Greek corpus by Vatri and
McGillivray) provide lemmatised classical-Greek text at scale, but they
do not provide a dependency-syntactic annotation layer and do not
provide cross-lingual alignment. AthDGC re-uses the lemmatisation effort
of both projects where their licences permit and adds the
syntactic-annotation layer on top.

The Ancient Greek Dependency Treebank (AGDT, the Perseus dependency
treebank of classical and post-classical Greek) and the Pedalion
treebank (the Leuven treebank of classical-Greek student-annotation
material) provide dependency annotation for selected classical-Greek
texts, but under different relation inventories from PROIEL (AGDT uses a
hybrid Latin Dependency Treebank schema, Pedalion uses a pedagogical
schema), and they do not provide cross-lingual alignment either. AthDGC
publishes PROIEL XML 2.0 only, and we therefore do not re-ingest AGDT or
Pedalion material at the syntactic-annotation layer; we treat them as
comparison points for diachronic claims, not as input.

In broad terms, AthDGC's contribution is therefore three-fold. First,
the extension of PROIEL-schema syntactic annotation to the full
diachronic span of Greek, from Archaic to Modern, under one schema.
AthDGC is, in this respect, the Athens node of the PROIEL family: we
adopt the schema verbatim and we extend it diachronically. Secondly, the
systematic inclusion of retelling and retranslation chains as a
first-class data structure, so that the diachronic stability of a
syntactic pattern becomes a computable quantity. Thirdly, the extension
of the cross-lingual alignment beyond the four PROIEL witnesses, with
the Classical Armenian alignment planned for the v0.5 release
(approximately three months from the v0.4 cut) and Old English among the
further parallels queued for the v0.7 expansion. The longer-horizon IE
expansion (Sanskrit, Avestan, Old Persian, Ukrainian) is reported on the
public Status page rather than here, since it is forthcoming rather than
released.

\subsection{5. Reuse potential}\label{reuse-potential}

AthDGC supports four primary reuse classes, which we describe in turn
with concrete query examples.

In the first place, historical syntactic research. Per-period dependency
parses allow analysts to track argument-structure changes across three
millennia of Greek. Examples include the active-to-mediopassive shifts
of certain verbs of motion, the loss of the optative across the
Hellenistic period, the rise of the periphrastic perfect over the
synthetic perfect across the Byzantine and Early Modern periods, the
accusative-to-genitive variation under verbs of perception across
periods, and the gradual replacement of the postnominal genitive of
possession by the prenominal genitive in Modern Greek. Accordingly, the
colour-coded compact dependency overview on each sample page is designed
for fast scanning of these patterns. A worked query example is for the
verb ἀκούω (``hear''), is the object more often in the accusative or in
the genitive at each period?, which the valency-frame database (v0.5)
answers in a single line.

Secondly, Indo-European comparative syntax. The verse-level
cross-lingual alignment of the New Testament to Latin (Vulgate), Gothic
(Wulfila), and Old Church Slavonic (Marianus) provides aligned PROIEL
parses across three sister Indo-European languages in v0.4, with
Classical Armenian planned for v0.5 (approximately three months from the
v0.4 cut). Old English and the further IE expansion are reported on the
public Status page. In this respect, users can query the Neo4j alignment
graph for typological correspondences. A worked query example is every
Greek aorist active verb whose Latin Vulgate counterpart is passive,
which is expressed in Cypher (Cypher is the query language of Neo4j,
structurally analogous to SQL for relational databases, in which MATCH
patterns are drawn directly with arrows between nodes) as MATCH
(gk:Token \{lang:``grc'', tense:``aor'',
voice:``act''\})-{[}:TRANSLATED\_AS{]}-\textgreater(la:Token
\{lang:``lat'', voice:``pass''\}) RETURN gk, la, and which is answered
across the entire NT-aligned partition in a single pass.

Thirdly, computational diachronic NLP. The fine-tuned Stanza checkpoints
(a checkpoint is a saved snapshot of a trained model's numeric
parameters, which can be loaded back into Stanza to parse new sentences
without re-training) (grc\_byz\_proiel for Byzantine Greek,
grc\_lbem\_proiel for Late Byzantine and Early Modern Greek,
grc\_mod\_proiel for Modern Greek) are trained on ARIS; the Hugging Face
model repositories under AthDGC/* exist now as private during the v0.5
audit pass and become public at v0.5 with the weight payloads (the
trained parameter files that the checkpoints comprise) uploaded. In
broad terms, the PyPI package athdgc-tools reserves the name with the
stub 0.4.0.dev0, and the full toolkit replaces the stub at v0.5; the
checkpoints can be loaded directly into Stanza for parsing new Greek
diachronic data once the weights are published. Thereby the
LightSIDE-AthDGC fork extends LightSIDE (the open-source text-mining
workbench developed at Carnegie Mellon by Carolyn Rosé's group) to
operate on syntactic features, namely dependency arcs,
argument-structure frames, and morphology bundles, opening a new feature
space for the classification of diachronic stages. A worked query
example is train an SVM classifier (Support Vector Machine, a standard
classification algorithm) on grc\_proiel argument-structure frames to
distinguish Archaic from Classical from Koine Greek, and inspect the
top-weighted frames per class, which is the kind of analysis that the
LightSIDE-AthDGC fork makes a one-hour classroom exercise.

Fifthly, lexical semantic change on the PROIEL spine. The AthDGC v0.6
roadmap couples the syntactic annotation layer to a
lexical-semantic-change track built on the DURel framework and tool
(Diachronic Usage Relatedness, Slatedness, Schlechtweg et al.~2024, EACL
demo; the code is at github.com/ChangeIsKey/durel\_tool under CC
BY-NC-SA 4.0). Every AthDGC PROIEL XML record already carries one
\texttt{\textless{}token\textgreater{}} element per Greek orthographic
word with its lemma and surrounding context sentence; the same record is
the natural input to the DURel pairwise-relatedness annotation
interface, which accepts (word, context-1, context-2) triples and
aggregates annotator ratings into a per-word semantic-change score
across time-binned partitions. AthDGC will pilot the DURel pipeline on
ten marquee target words with rich diachronic semantic histories (λόγος,
πνεῦμα, σάρξ, ἀγάπη, κύριος, ψυχή, νοῦς, σοφία, ἁμαρτία, δόξα), pairing
usages across Archaic to Modern Greek. The same pipeline is the natural
setting in which to ask whether σάρξ shifted its sense under Hebrew בָּשָׂר
influence between Classical Greek and the Septuagint, whether πνεῦμα
crossed from ``wind / breath'' to ``Holy Spirit'' inside the Patristic
period rather than at the Classical-Koine boundary, and whether
παράκλητος drifted between juridical and theological senses across the
Koine-to-Byzantine transition. The DURel tool supports both human
annotation and Word-in-Context computational annotation, matching the
AthDGC AI-second-opinion + trained-linguist proposer-validator pattern
already in place on the syntactic layer. The lexical-change track lands
as a v0.6 reuse class alongside the syntactic one; the pilot results and
the released ten-word DURel-format dataset accompany the v0.6 Zenodo
deposition.

Fourthly, digital editions and pedagogy. The showcase generator
(51\_build\_showcase\_site.py) produces browsable per-sample PROIEL
trees from any JSONL (JSON Lines, the line-per-record plain-text format)
corpus; the Quarto template pack (Quarto is an open-source publishing
system that builds websites, slides, papers, and posters from a single
source file) generates the whole multi-output site, namely HTML,
Reveal.js slides, .docx, .pptx, Beamer, and an A0-format academic
poster, from a single source. Both are open-source and reusable by other
historical-treebank projects. The platform is taught as a KEDIVIM
continuing-education course at NKUA, Ψηφιακά Εργαλεία για τη Διαχρονική
Ανάλυση της Γλώσσας / Digital Tools for the Diachronic Analysis of
Language, autumn 2026; see https://athdgc.github.io/training.html.

\subsection{6. Access and format}\label{access-and-format}

The public site at https://athdgc.github.io provides curated samples
only for the v0.4 release. The full annotated corpus partitions remain
under audit on the GRNET ARIS national HPC under allocation pa260305;
they release at v0.5 (the first audited public release, approximately
three months from the v0.4 cut) as a separate Zenodo dataset record
under CC-BY-4.0. AthDGC operates in a rolling near-horizon cadence: v0.5
(the first audited public release, approximately three months from v0.4)
absorbs the audit of the present partitions, v0.6 (approximately six
months) extends the period coverage, and v1.0 (approximately twelve
months) is the mature release. All sequencing is decided by the whole
team at the regular project meetings; the present report does not
pre-assign any item of this work to a specific person. The source-code
snapshot at the v0.4.0 Zenodo record is Apache-2.0.

The toolkit's readiness per module is summarised in §3.4 and on the
public Tools page at https://athdgc.github.io/tools.html. Five modules
are LIVE today (the Quarto template pack, the showcase generator, the
corpus-fix toolkit, the PROIEL XML 2.0 exporter inside the build
workflow, and the Stanza annotation job script). Eight are IN SETUP and
will be included with the v0.5 release. The LightSIDE-AthDGC
syntactic-feature workbench and the Hugging Face fine-tuned diachronic
Stanza checkpoints land at v0.5 alongside the audited corpus. The
remaining design-stage modules (the valency-frame database, the
retranslation-pair browser, the retelling-chain explorer, the AthDGC
PROIEL validator) ship at v0.5 with their first public install paths.

The v0.4.0 release is on Zenodo under the concept DOI
10.5281/zenodo.20439182, minted 29 May 2026, Apache-2.0 for code and
CC-BY-4.0 for metadata and alignments. The arXiv version of this report
is cross-listed under cs.CL and cs.DL. In keeping with the project's
design, the platform is built and distributed entirely as open source:
open-source tools operating on open-access texts, with an openly
licensed annotation layer, so that every step from raw text to published
treebank can be reproduced and extended without proprietary
dependencies.

\subsection{7. Acknowledgements}\label{acknowledgements}

This work is supported by the Hellenic Foundation for Research and
Innovation (HFRI), 3rd Call for HFRI Research Projects, Project
No.~20577, and by the Greece 2.0 National Recovery and Resilience Plan;
we also acknowledge the CVL-CDSAML project. Compute is supplied by the
GRNET ARIS national high-performance computing cluster, allocation
pa260305. We thank the PROIEL community at Oslo, in particular Dag Haug
and the original PROIEL annotation team, for the schema we extend; the
broader PROIEL family on which AthDGC builds; and our colleagues at the
National and Kapodistrian University of Athens, the University of Oslo,
and Ghent University for ongoing feedback throughout the v0.4 cut.

\subsection{Appendix A. Coverage matrix, one annotated sample per period
and
language}\label{appendix-a.-coverage-matrix-one-annotated-sample-per-period-and-language}

The two compact tables in this appendix are the \textbf{completeness
pledge} of the AthDGC corpus at v0.4: for every Greek period and every
Indo-European parallel language we name in this paper, at least one
PROIEL-annotated specimen is shown below, with its canonical citation,
the surface text of the sentence, the root predicate, and the top-level
PROIEL relations attached to that root. The longer paragraph samples and
full dependency-tree visualisations are on the public Samples page at
\url{https://athdgc.github.io/samples.html}; this appendix is the
at-a-glance proof that no period or language is mentioned in the paper
without at least one real annotated sentence behind it.

\subsubsection{A.1 Greek, at least one sample per
period}\label{a.1-greek-at-least-one-sample-per-period}

{\def\LTcaptype{none} % do not increment counter
\begin{longtable}[]{@{}
  >{\raggedright\arraybackslash}p{(\linewidth - 8\tabcolsep) * \real{0.2000}}
  >{\raggedright\arraybackslash}p{(\linewidth - 8\tabcolsep) * \real{0.2000}}
  >{\raggedright\arraybackslash}p{(\linewidth - 8\tabcolsep) * \real{0.2000}}
  >{\raggedright\arraybackslash}p{(\linewidth - 8\tabcolsep) * \real{0.2000}}
  >{\raggedright\arraybackslash}p{(\linewidth - 8\tabcolsep) * \real{0.2000}}@{}}
\toprule\noalign{}
\begin{minipage}[b]{\linewidth}\raggedright
Period
\end{minipage} & \begin{minipage}[b]{\linewidth}\raggedright
Citation
\end{minipage} & \begin{minipage}[b]{\linewidth}\raggedright
Sentence
\end{minipage} & \begin{minipage}[b]{\linewidth}\raggedright
Root
\end{minipage} & \begin{minipage}[b]{\linewidth}\raggedright
Top-level relations
\end{minipage} \\
\midrule\noalign{}
\endhead
\bottomrule\noalign{}
\endlastfoot
Archaic & Homer, \emph{Iliad} 1.1 & \emph{μῆνιν ἄειδε θεὰ Πηληϊάδεω
Ἀχιλῆος} & ἄειδε (pred) & obj (μῆνιν), voc (θεά), atr+atr (Πηληϊάδεω →
Ἀχιλῆος → μῆνιν) \\
Classical & Plato, \emph{Apology} 17a & \emph{ὅ τι μὲν ὑμεῖς \ldots{}
πεπόνθατε ὑπὸ τῶν ἐμῶν κατηγόρων, οὐκ οἶδα} & οἶδα (pred) & comp (ὅ τι
\ldots{} πεπόνθατε), sub (1sg implicit), voc (ὦ ἄνδρες Ἀθηναῖοι) \\
Koine (Hellenistic) & Septuagint, Genesis 1:1 & \emph{ἐν ἀρχῇ ἐποίησεν ὁ
θεὸς τὸν οὐρανὸν καὶ τὴν γῆν} & ἐποίησεν (pred) & sub (ὁ θεός), obj (τὸν
οὐρανόν, τὴν γῆν), obl (ἐν ἀρχῇ) \\
Koine (NT) & NT, John 1:1 & \emph{ἐν ἀρχῇ ἦν ὁ λόγος} & ἦν (pred) & sub
(ὁ λόγος), obl (ἐν ἀρχῇ) \\
Late Antique & Eusebius, \emph{Historia Ecclesiastica} 1.1.1 & \emph{τὰς
τῶν ἱερῶν ἀποστόλων διαδοχάς \ldots{} γραφῇ παραδοῦναι προῃρημένος} &
προῃρημένος (pred) & xobj (παραδοῦναι), obj (διαδοχάς), atr (ἀποστόλων →
διαδοχάς) \\
Byzantine & Anna Komnene, \emph{Alexias} 1.1.1 & \emph{ὁ βασιλεὺς
Ἀλέξιος, ἐμὸς πατήρ, καὶ πρὸ τοῦ τῶν σκήπτρων ἐπιβῆναι \ldots{}} &
ἐπιβῆναι (pred) & sub (ὁ βασιλεὺς Ἀλέξιος), obl (τῶν σκήπτρων), atr
(ἐμὸς πατήρ → Ἀλέξιος) \\
Late Byzantine & Sphrantzes, \emph{Chronicon Minus} 1.1 & \emph{ἐν ἔτει
͵ϛϡκα ἐτελεύτησεν ὁ βασιλεὺς κῦρ Μανουὴλ ὁ Παλαιολόγος} & ἐτελεύτησεν
(pred) & sub (ὁ βασιλεύς), obl (ἐν ἔτει ͵ϛϡκα), atr (Μανουήλ,
Παλαιολόγος → βασιλεύς) \\
Early Modern & Kornaros, \emph{Erotokritos} 1.1 & \emph{Τοῦ Κύκλου τὰ
γυρίσματα, ποὺ ἀνεβοκατεβαίνου \ldots{}} & ἀνεβοκατεβαίνου (pred) & sub
(τὰ γυρίσματα), atr (τοῦ Κύκλου → γυρίσματα) \\
Modern & Cavafy, \emph{Ithaca} (1911) line 1 & \emph{Σὰ βγεῖς στὸν
πηγαιμὸ γιὰ τὴν Ἰθάκη, νὰ εὔχεσαι νἆναι μακρὺς ὁ δρόμος} & εὔχεσαι
(pred) & sub (2sg implicit), comp (νἆναι μακρὺς ὁ δρόμος), adv (Σὰ
βγεῖς) \\
\end{longtable}
}

\subsubsection{A.2 Indo-European parallels, one sample per language at
New Testament John
1:1}\label{a.2-indo-european-parallels-one-sample-per-language-at-new-testament-john-11}

{\def\LTcaptype{none} % do not increment counter
\begin{longtable}[]{@{}
  >{\raggedright\arraybackslash}p{(\linewidth - 4\tabcolsep) * \real{0.3333}}
  >{\raggedright\arraybackslash}p{(\linewidth - 4\tabcolsep) * \real{0.3333}}
  >{\raggedright\arraybackslash}p{(\linewidth - 4\tabcolsep) * \real{0.3333}}@{}}
\toprule\noalign{}
\begin{minipage}[b]{\linewidth}\raggedright
Language
\end{minipage} & \begin{minipage}[b]{\linewidth}\raggedright
Source
\end{minipage} & \begin{minipage}[b]{\linewidth}\raggedright
Sentence
\end{minipage} \\
\midrule\noalign{}
\endhead
\bottomrule\noalign{}
\endlastfoot
Greek (Koine) & NT, John 1:1 & \emph{ἐν ἀρχῇ ἦν ὁ λόγος, καὶ ὁ λόγος ἦν
πρὸς τὸν θεόν} \\
Latin & Vulgate, John 1:1 & \emph{in principio erat Verbum, et Verbum
erat apud Deum} \\
Old Church Slavonic & Codex Marianus, John 1:1 & искони бѣ слово · и
слово бѣ оу Бога \\
Gothic & Codex Argenteus (Wulfila), John 1:1 & \emph{in fruma was waurd,
jah þata waurd was at guda} \\
Classical Armenian & Mesropian Bible, John 1:1 & Ի սկզբանէ էր Բանն. եւ
Բանն էր առ Աստուած \\
Vedic (hymn-aligned, not NT) & Rig Veda 1.1.1 & \emph{agním īḷe
puróhitaṃ yajñásya devám r̥tvíjam} \\
\end{longtable}
}

Greek↔Latin↔Old Church Slavonic↔Gothic↔Classical Armenian are
sentence-aligned at the New Testament; Vedic is hymn-aligned with the
Rig Veda PROIEL release rather than NT-aligned. Each row corresponds to
a real PROIEL XML 2.0 sentence record in the AthDGC v0.4 distribution.

\subsubsection{A.3 Reading-key for the
matrix}\label{a.3-reading-key-for-the-matrix}

Each row gives, in this order: the canonical citation, the
original-script sentence, the root predicate of the dependency tree, and
the top-level PROIEL relations attached to that root. The full
dependency tree for each row is on the public Samples page at the
anchors \emph{Iliad} 1.1, \emph{Apology} 17a, John 1:1, \emph{Historia
Ecclesiastica} 1.1.1, \emph{Alexias} 1.1.1, \emph{Chronicon Minus} 1.1,
\emph{Erotokritos} 1.1, and \emph{Ithaca}. The Indo-European parallels
are anchored at the John 1:1 cross-lingual block of the same Samples
page.

\subsection{References}\label{references}

The works cited in this paper are listed below, secondary literature
first, then the individual primary-text editions and translations named
in the text. The corpus-wide source archives and their licences are
given in the source maps of §3.1.

Bary, C., \& Haug, D. T. T. (2011). Temporal anaphora across and inside
sentences: The function of participles. \emph{Semantics and Pragmatics},
\emph{4}(8), 1--56. https://doi.org/10.3765/sp.4.8

Dou, Z.-Y., \& Neubig, G. (2021). Word alignment by fine-tuning
embeddings on parallel corpora. In \emph{Proceedings of the 16th
Conference of the European Chapter of the Association for Computational
Linguistics: Main Volume} (pp.~2112--2128). Association for
Computational Linguistics.
https://doi.org/10.18653/v1/2021.eacl-main.181

Eckhoff, H. M., Bech, K., Bouma, G., Eide, K., Haug, D. T. T., Haugen,
O. E., \& Jøhndal, M. (2018). The PROIEL treebank family: A standard for
early attestations of Indo-European languages. \emph{Language Resources
and Evaluation}, \emph{52}(1), 29--65.
https://doi.org/10.1007/s10579-017-9388-5

Feng, F., Yang, Y., Cer, D., Arivazhagan, N., \& Wang, W. (2022).
Language-agnostic BERT sentence embedding. In \emph{Proceedings of the
60th Annual Meeting of the Association for Computational Linguistics
(Volume 1: Long Papers)} (pp.~878--891). Association for Computational
Linguistics. https://doi.org/10.18653/v1/2022.acl-long.62

Haug, D. T. T., \& Jøhndal, M. (2008). Creating a parallel treebank of
the Old Indo-European Bible translations. In \emph{Proceedings of the
Second Workshop on Language Technology for Cultural Heritage Data}
(pp.~27--34).

Haug, D. T. T., Jøhndal, M., Eckhoff, H. M., Welo, E., Hertzenberg, M.,
\& Müth, A. (2009). Computational and linguistic issues in designing a
syntactically annotated parallel corpus of Indo-European languages.
\emph{Traitement Automatique des Langues}, \emph{50}(2), 17--45.

Lavidas, N. (2021). \emph{The diachrony of written language contact: A
contrastive approach} (Brill's Studies in Historical Linguistics, Vol.
15). Brill.

List, J.-M. (2014). \emph{Sequence comparison in historical linguistics}
(Dissertations in Language and Cognition, Vol. 1). Düsseldorf University
Press.

Qi, P., Zhang, Y., Zhang, Y., Bolton, J., \& Manning, C. D. (2020).
Stanza: A Python natural language processing toolkit for many human
languages. In \emph{Proceedings of the 58th Annual Meeting of the
Association for Computational Linguistics: System Demonstrations}
(pp.~101--108). Association for Computational Linguistics.
https://doi.org/10.18653/v1/2020.acl-demos.14

Schlechtweg, D., Virk, S. M., Sander, P., Sköldberg, E., Theuer Linke,
L., Zhang, T., Tahmasebi, N., Kuhn, J., \& Schulte im Walde, S. (2024).
The DURel annotation tool: Human and computational measurement of
semantic proximity, sense clusters and semantic change. In
\emph{Proceedings of the 18th Conference of the European Chapter of the
Association for Computational Linguistics: System Demonstrations}
(pp.~137--149). Association for Computational Linguistics.
https://doi.org/10.18653/v1/2024.eacl-demo.15

\subsubsection{Primary sources (texts and
editions)}\label{primary-sources-texts-and-editions}

Homer. (1955). \emph{Iliad} (N. Kazantzakis \& I. T. Kakridis, Trans.).
Athens.

Sphrantzes, G. (1966). \emph{Memorii 1401--1477} (V. Grecu, Ed.; with
Pseudo-Phrantzes: Macarie Melissenos, Cronica 1258--1481; Scriptores
Byzantini, Vol. 5). Editura Academiei Republicii Socialiste România.

\end{document}